\documentclass[pmlr]{jmlr}
\pdfoutput=1

\usepackage{longtable}
\usepackage{mathabx}

\usepackage{booktabs}
\usepackage[load-configurations=version-1]{siunitx} 

\usepackage{multirow}

\jmlrvolume{1} 
\jmlryear{2019}
\jmlrworkshop{NeurIPS2019 Disentanglement Challenge}

\title[Improved Disentanglement through Learned Aggregation of Feature Maps]{NeurIPS 2019 Disentanglement Challenge: Improved Disentanglement through Learned Aggregation of Convolutional Feature Maps}

  \author{\Name{Maximilian Seitzer} \Email{contact@max-seitzer.de}\\
  \Name{Andreas Foltyn} \Email{andreas.foltyn@iis.fraunhofer.de}\\
  \Name{Felix P. Kemeth} \Email{felix.kemeth@iis.fraunhofer.de}\\
  \addr Fraunhofer Institute for Integrated Circuits IIS, Erlangen, Germany}

\begin{document}

\maketitle

\begin{abstract}
This report to our stage 2 submission to the NeurIPS 2019 disentanglement challenge presents a simple image preprocessing method for learning disentangled latent factors. 
We propose to train a variational autoencoder on regionally aggregated feature maps obtained from networks pretrained on the ImageNet database, utilizing the implicit inductive bias contained in those features for disentanglement.
This bias can be further enhanced by explicitly fine-tuning the feature maps on auxiliary tasks useful for the challenge, such as angle, position estimation, or color classification.
Our approach achieved the 2nd place in stage 2 of the challenge~\citep{AicrowdChallenge2019}.
Code is available at \url{https://github.com/mseitzer/neurips2019-disentanglement-challenge}.

\end{abstract}

\section{Introduction}
\label{sec:intro}

Fully unsupervised methods are unable to learn disentangled representations without introducing further assumptions in the form of inductive biases on model and data~\citep{Locatello2018ChallengingCA}.
In our challenge submission, we utilize the implicit inductive bias contained in models pretrained on the ImageNet database~\citep{Russakovsky2014ImageNetLS}, and enhance it by finetuning such models on challenge-relevant auxiliary tasks such as angle, position estimation, or color classification.
In particular, our submission for stage 2 builds on our submission from stage 1~\citep{Seitzer2020stage1}, in which we employed pretrained CNNs to extract convolutional feature maps as a preprocessing step before training a VAE~\citep{Kingma2013AutoEncodingVB}.
Although this approach already yielded good disentanglement scores, we identified two weaknesses with the feature vectors extracted this way. 
First, the feature extraction network is trained on ImageNet, which is rather dissimilar to the \emph{MPI3d} dataset~\citep{Gondal2019OnTT} used in the challenge. 
Secondly, the feature aggregation mechanism was chosen ad-hoc and likely does not retain all information needed for disentanglement. 
We attempt to alleviate these issues by finetuning the feature extraction network as well as learning the aggregation of feature maps from data by using the labels of the simulation datasets \emph{MPI3d-toy} and \emph{MPI3d-realistic}.

\section{Method}
Our method consists of the following three steps: (1) supervised finetuning of the feature extraction CNN (section \ref{subsec:finetuning}), 
(2) extracting a feature vector from each image in the dataset using the finetuned network (section \ref{subsec:extraction_and_aggregation}), (3) training a VAE to reconstruct the feature vectors and disentangle the latent factors of variation (section \ref{subsec:training}).

\subsection{Finetuning the Feature Extraction Network}
\label{subsec:finetuning}

In this step, we finetune the feature extraction network offline (before submission to the evaluation server). 
The goal is to adapt the network such that it produces aggregated feature vectors that retain the information needed to disentangle the latent factors of the \emph{MPI3d-real} dataset.
In particular, the network is finetuned by learning to predict the value of each latent factor from the aggregated feature vector of an image.
To this end, we use the simulation datasets \emph{MPI3d-toy} and \emph{MPI3d-realistic}\footnote{Pretraining using \emph{any} data was explicitly stated to be allowed by the challenge organizers.}, namely the images as inputs and the labels as supervised classification targets.

For the feature extraction network, we use the VGG19-BN architecture \citep{Simonyan2014VeryDC} of the \texttt{torchvision} package.
The input images are standardized using mean and variance across each channel computed from the ImageNet dataset.
We use the output feature maps of the last layer before the final average pooling (dimensionality $512 \times 2 \times 2$) as the input to a feature aggregation module which reduces the feature map to a $512$-dimensional vector\footnote{This reduction to aggregated feature vectors instead of directly using feature maps was necessitated by the memory requirements of the challenge.}.
This aggregation module consists of three convolution layers with $1024, 2048, 512$ feature maps and kernel sizes $1, 2, 1$ respectively.
Each layer is followed by batch normalization and ReLU activation.
We also employ layerwise dropout with rate $0.1$ before each convolution layer.
Finally, the aggregated feature vector is $\ell 2$-normalized, which was empirically found to be important for the resulting disentanglement performance.
Then, for each latent factor, we add a linear classification layer computing the logits of each class from the aggregated feature vector.
These linear layers are discarded after this step.

We use both \emph{MPI3d-toy} and \emph{MPI3d-realistic} for training to push the network to learn features that identify the latent factors in a robust way, regardless of details such as reflections or specific textures. 
In particular, we use a random split of 80\% of each dataset as the training set, and the remaining samples as a validation set.
VGG19-BN is initialized with a set of weights resulting from ImageNet training\footnote{\url{https://download.pytorch.org/models/vgg19_bn-c79401a0.pth}}, and the aggregation module and linear layers were randomly initialized using uniform He initialization \citep{He2015DelvingDI}.
The network is trained for 5 epochs using the RAdam optimizer~\citep{Liu2019Radam} with learning rate $0.001$, $\beta_0=0.999$, $\beta_1=0.9$, a batch size of $512$ and a weight decay of $0.01$.
We use a multi-task classification loss consisting of the sum of cross entropies between the prediction and the ground truth of each latent factor.
After training, the classification accuracy on the validation set is around 98\% for the two degrees of freedom of the robot arm, and around 99.9\% for the remaining latent factors.

\subsection{Feature Map Extraction and Aggregation}
\label{subsec:extraction_and_aggregation}

In this step, we use the finetuned feature extraction network to produce a set of aggregated feature vectors. 
We simply run the network detailed in the previous step on each image of the dataset and store the aggregated $512$-dimensional vectors in memory.
Again, inputs to the feature extractor are standardized such that mean and variance across each channel correspond to the respective ones from the ImageNet dataset.

\subsection{VAE Training}
\label{subsec:training}

Finally, we train a standard $\beta$-VAE~\citep{Higgins2017betaVAELB} on the set of aggregated feature vectors resulting from the previous step.
The encoder network consists of a single fully-connected layers with $4096$ neurons, followed by two fully-connected layers parametrizing $C=18$ means and log variances of a normal distribution $\mathcal{N} \left(\vec{\mu}(\vec{x}), \vec{\sigma}^2(\vec{x}) \right)$ used as the approximate posterior $q\left(\vec{z} \mid \vec{x}\right)$.
The number of latent factors was experimentally determined.
The decoder network consists of four fully-connected layers with $4096$ neurons each, followed by a fully-connected layer parametrizing the means of a normal distribution $\mathcal{N} \left(\vec{\hat{\mu}}(\vec{z}), \vec{I} \right)$ used as the conditional likelihood $p\left(\vec{x} \mid \vec{z}\right)$.
The mean is constrained to range $(0, 1)$ using the sigmoid activation.
All fully-connected layers but the final ones use batch normalization and are followed by ReLU activation functions.
We use orthogonal initialization~\citep{Saxe2013Orth} for all layers and assume a factorized standard normal distribution $\mathcal{N}\left(\vec{0}, \vec{I}\right)$ as the prior $p\left(\vec{z}\right)$ on the latent variables.

For optimization, we use the RAdam optimizer~\citep{Liu2019Radam} with a learning rate of $0.001$, $\beta_0=0.999$, $\beta_1=0.9$ and a batch size of $B=256$.
The VAE is trained for $N=120$ epochs by maximizing the evidence lower bound, which is equivalent to minimizing 

$$\frac{1}{B} \sum_{i=1}^{512} \left(\hat{\mu}_i - x_i\right)^2 - 0.5 \frac{\beta}{BC} \sum_{j=1}^C 1 + \log(\sigma_j^2) - \mu_j^2 - \sigma_j^2$$

where $\beta$ is a hyperparameter to balance the MSE reconstruction and the KLD penalty term.
As the scale of the KLD term depends on the numbers of latent factors $C$, we normalize it by $C$ such that $\beta$ can be varied independently of $C$.
It can be harmful to start training with too much weight on the KLD term~\citep{Bowman2015GeneratingSF}.
Therefore, we use the following cosine schedule to smoothly anneal $\beta$ from $\beta_\text{start}=0.005$ to $\beta_\text{end}=0.4$ over the course of training:

$$\beta(t) = \begin{cases}
\beta_\text{start} &\mbox{for } t < t_\text{start} \\
\beta_\text{end} - \frac{1}{2} \left(\beta_\text{end} - \beta_\text{start} \right) \left(1 + \cos \pi \frac{t - t_\text{start}}{t_\text{end} - t_\text{start}} \right) &\mbox{for } t_\text{start} \leq t \leq t_\text{end}\\
\beta_{\text{end}} &\mbox{for } t > t_\text{end}
\end{cases}$$

where $\beta(t)$ is the value for $\beta$ in training episode $t \in \{0, \dots, N - 1\}$, and annealing runs from epoch $t_\text{start}=10$ to epoch $t_\text{end}=79$.
This schedule lets the model initially learn to reconstruct the data well and only then puts pressure on the latent variables to be factorized which we found to considerably improve performance.

\section{Discussion}

Our approach was able to obtain the second place in stage 2 of the competition.
Notably, compared to our stage 1 approach, our stage 2 approach leads to a large improvement on the FactorVAE~ \citep{Kim2018DisentanglingBF}, and DCI~\citep{Eastwood2018AFF} metrics.
On the public leaderboard, our best submission achieves the first rank on these metrics, with a large gap to the second-placed entry.
See appendix \ref{apd:results} for further discussion of the results.

Unsurprisingly, introducing prior knowledge simplifies the disentanglement task considerably, which is reflected in the improved scores. 
To do so, our approach makes use of task-specific supervision obtained from simulation, which restricts its applicability.
Nevertheless, it constitutes a demonstration that this type of supervision can transfer to better disentanglement on real world data, which was one of the goals of the challenge.

\section*{Acknowledgements}
This work was supported by the Bavarian Ministry of Economic Affairs, Regional Development and Energy through the Center for Analytics - Data - Applications (ADA-Center) within the framework of "BAYERN DIGITAL II"  (20-3410-2-9-8).

\bibliography{references}

\appendix

\section{Discussion of Results on Leaderboard Results}
\label{apd:results}
We summarize the results of our best submissions on the public and private leaderboards in table \ref{tab:results}. 
The private leaderboard of this challenge stage used a dataset of real images, but with objects more difficult to recognize than the objects in the dataset of the public leaderboard.
An exact description of the types of objects used for this test dataset was not yet released at the time of writing to the best of our knowledge.
On this more difficult dataset, our approach achieves the first rank on the FactorVAE~ \citep{Kim2018DisentanglingBF} and SAP~\citep{Kumar2017VariationalIO} metrics, with a particularly large difference of $0.24$ to the second ranked entry for FactorVAE.
Compared to the easier dataset of the public leaderboard, all metrics drop, sometimes strongly (e.\,g. $0.22$ for DCI~\citep{Eastwood2018AFF}).
This could stem from the fact that this more challenging dataset uses different types of objects than the ones which were included in the supervised pretraining.

On the public leaderboard (i.\,e. on \emph{MPI3D-real}), our method achieves the first rank on FactorVAE and DCI. 
For both metrics, there is a large absolute difference to the second ranked entry, namely $0.37$ for FactorVAE and $0.26$ for DCI. 
For SAP, our method is almost tied with the first ranking entry, with $0.01$ absolute difference.
For MIG~\citep{Chen2018IsolatingSO} and IRS~\citep{Suter2019RobustlyDC}, our method falls behind the best method, with an absolute distance of $0.08$ and $0.13$ respectively.

Compared to our stage 1 submission which does not use supervised finetuning, metrics for which our approach was already good (FactorVAE, DCI and SAP), became even better, while other metrics for which our approach performed subpar stayed the same (MIG) or even became worse (IRS). 
It seems that adding supervised finetuning to our pretrained features approach enhances the already existing strengths and weaknesses.
That being said, it is known that the results of VAE-based disentanglement methods are highly sensitive to the hyperparameters and even random seeds used~\citep{Locatello2018ChallengingCA}.
Thus a more detailed investigation is needed to draw any conclusions, which was out of scope for this report.

\begin{table}[t]
\floatconts
  {tab:results}%
  {\caption{Summary of scores and ranks of our best submissions on the private and public leaderboards at the end of stage 2. 
  For comparison, we also include the private leaderboard scores of our best stage 1 submission.
  Note that our best result on the public leaderboard uses slighly different hyperparameters than the ones described before. 
  We list them in appendix \ref{apd:hparams}.
  }}%
  {\begin{tabular}{llccccccc}
  \toprule
  & \bfseries Dataset & \bfseries FactorVAE & \bfseries DCI & \bfseries SAP & \bfseries IRS & \bfseries MIG & \\
  \midrule
Private Score & MPI3d-real & 0.893 & 0.589 & 0.192 & 0.447 & 0.268 & $\sum 2.389$ \\
Rank (of 11) & (difficult) & 1 & 2 & 1 & 11 & 3 & $\diameter \, 3.6$ \\ 
  \midrule
Public Score & \multirow{2}{*}{MPI3d-real} & 0.992 & 0.809 & 0.223 & 0.547 & 0.297 & $\sum 2.868$ \\
Rank (of 11) & & 1 & 1 & 2 & 11 & 3 & $\diameter \, 3.6$ \\ 
  \midrule
Score Stage 1 & \multirow{2}{*}{MPI3d-real} & 0.792 & 0.527 & 0.166 & 0.623 & 0.292 & $\sum 2.400$ \\
Rank (of 35) & & 1 & 2 & 2 & 21 & 3 & $\diameter \, 5.8$ \\ 
  \bottomrule
  \end{tabular}}
\end{table}

\section{Hyperparameters for Best Result on Public Leaderboard}
\label{apd:hparams}

The challenge leaderboard lists only the best submissions for each dataset, which is why the best submission on the public leaderboard has slighly different hyperparameters than the best one on the private dataset.
We list the hyperparameters different from the ones described in the main text here:

\begin{itemize}
 \itemsep0em 
 \item Encoder: four layers with $1024$ neurons each
 \item Decoder: four layers with $1024$ neurons each
 \item Latent dimensions: $C=16$
 \item Training time: $N=100$ epochs
 \item Beta annealing: from $\beta_\text{start}=0.001$ to $\beta_\text{end}=0.4$ over epochs $t_\text{start}=10$ to $t_\text{end}=49$
\end{itemize}

\end{document}